\documentclass[11pt,a4paper]{article}
\usepackage[hyperref]{emnlp2018}
\usepackage{times}
\usepackage{latexsym}

\usepackage{url}

\usepackage{microtype}      

\usepackage{multirow}
\usepackage{graphicx}
\usepackage{amsmath}

\usepackage{arydshln}

\title{A Study of Reinforcement Learning for Neural Machine Translation}

\author{Lijun Wu$^{1}$, Fei Tian$^2$, Tao Qin$^2$, Jianhuang Lai$^{1}$ \and Tie-Yan Liu$^2$\\
$^1$School of Data and Computer Science, Sun Yat-sen University\\
\quad$^2$Microsoft Research \\
wulijun3@mail2.sysu.edu.cn;\;stsljh@mail.sysu.edu.cn;\;\\
\{fetia,taoqin,tyliu\}@microsoft.com\;
}

\date{}

\begin{document}
\maketitle
\begin{abstract}
Recent studies have shown that reinforcement learning (RL) is an effective approach for improving the performance of neural machine translation (NMT) system.  However, due to its instability,  successfully RL training is challenging, especially in real-world systems where deep models and large datasets are leveraged. In this paper, taking several large-scale translation tasks as testbeds, we conduct a systematic study on how to train better NMT models using reinforcement learning. We provide a comprehensive comparison of several important factors (e.g., baseline reward, reward shaping) in RL training. Furthermore, to fill in the gap that it remains unclear whether RL is still beneficial when monolingual data is used,  we propose a new method to leverage RL to further boost the performance of NMT systems trained with source/target monolingual data. By integrating all our findings, we obtain competitive results on WMT14 English-German, WMT17 English-Chinese, and WMT17 Chinese-English translation tasks, especially setting a state-of-the-art performance on WMT17 Chinese-English translation task.

\end{abstract}

\section{Introduction}

Recently, neural machine translation (NMT) \cite{attention,hassan2018achieving,gnmt,he2017decoding,duallearning,Deliberation,wordatt,error_prop} has become more and more popular given its superior performance without the demand of heavily hand-crafted engineering efforts. It is usually trained to maximize the likelihood of each token in the target sentence, by taking the source sentence and the preceding (ground-truth) target tokens as inputs. Such training approach is referred as maximum likelihood estimation (MLE) \cite{mle}. 
Although easy to implement, the token-level objective function during training is inconsistent with sequence-level evaluation metrics such as BLEU \cite{bleu}.

To address the inconsistency issue, reinforcement learning (RL) methods have been adopted to optimize sequence-level objectives. For example, policy optimization methods such as REINFORCE \cite{mixer,monoRL} and actor-critic \cite{AC4NMT} are leveraged for sequence generation tasks including NMT. In machine translation community, a similar method is proposed with the name `minimum risk training' \cite{MRT}.  All these works demonstrate the effectiveness of RL techniques for NMT models~\cite{gnmt}.

However, effectively applying RL to real-world NMT systems has not been fulfilled by previous works. First, most of, if not all, previous works verified their methods based on shallow recurrent neural network (RNN) models. However, to obtain state-of-the-art (SOTA) performance, it is essential to leverage recently derived deep models \cite{fairseq, transformer}, which are much more powerful.

Second, it is not easy to make RL practically effective given quite a few widely acknowledged limitations of RL method \cite{drlhard} such as high variance of gradient estimation \cite{weaver2001optimal}, and objective instability \cite{mnih2013playing}. Therefore, several tricks are proposed in previous works. However, it remains unclear, and no agreement is achieved on how to use these tricks in machine translation. For example, baseline reward method~\cite{weaver2001optimal} is suggested in \cite{mixer,BanditNMT,gnmt} but not leveraged in~\cite{he2012maximum,MRT}.

Third, large-scale datasets, especially monolingual datasets are shown to significantly improve translation quality~\cite{backtranslation,duallearning} with MLE training, while it remains nearly empty on how to combine RL with monolingual data in NMT.

In this paper, we try to fulfill these gaps and study how to practically apply RL to obtain strong NMT systems with quite competitive, even state-of-the-art performance.
Several comprehensive studies are conducted on different aspects of RL training to figure out how to: 1)  set efficient rewards; 2) combine MLE and RL objectives with different weights, which aims to stabilize the training procedure; 3) reduce the variance of gradient estimation.

In addition, given the effectiveness of leveraging monolingual data in improving translation quality, we further propose a new method to combine the strength of both RL training and source/target monolingual data. To the best of our knowledge, this is the first work that tries to explore the power of monolingual data when training NMT model with RL method.

We obtain some useful findings through the experiments on WMT17 Chinese-English (Zh-En), WMT17 English-Chinese (En-Zh) and WMT14 English-German (En-De) translation tasks. For instance, multinomial sampling is better than beam search in reward computation, and the combination of RL and monolingual data significantly enhances the NMT model performance. Our main contributions are summarized as follows.

\begin{itemize}
\item We provide the first comprehensive study on different aspects of RL training, such as how to setup reward and baseline reward, on top of quite competitive NMT models.
\item We propose a new method that effectively leverages large-scale monolingual data, from both the source and target side,  when training NMT models with RL.
\item Combined with several of our findings and method, we obtain the SOTA translation quality on WMT17 Zh-En translation task, surpassing strong baseline (Transformer big model + back translation) by nearly 1.5 BLEU points. Furthermore, on WMT14 En-De and WMT17 En-Zh translation tasks, we can also obtain strong competitive results.
\end{itemize}

We hope that our studies and findings will benefit the community to better understand and leverage reinforcement learning for developing strong NMT models, especially in real-world scenarios faced with deep models and large amount of training data (including both parallel and monolingual data). Towards this end, we open source all our codes/dataset at \url{https://github.com/apeterswu/RL4NMT} to provide a clear recipe for performance reproduction.

\section{Background}

In this section, we first introduce the attention-based sequence-to-sequence learning framework for neural machine translation (NMT), and then introduce the basis of applying reinforcement learning to training NMT models.

\subsection{Neural Machine Translation}

Typical NMT models are based on the encoder-decoder framework with attention mechanism. The encoder first maps a source sentence $x=(x_1, x_2,...,x_n)$ to a set of continuous representations $z = (z_1, z_2,...,z_n)$. Given $z$, the decoder then generates a target sentence $y=(y_1, y_2,...,y_m)$ of word tokens one by one. At each decoding step $t$ of model training, the probability of generating a token $y_t$ is maximized conditioned on $x$ and $y_{<t}=(y_1,...,y_{t-1})$. Given $N$ training sentence pairs $\{x^i,y^i\}_{i=1}^N$, maximum likelihood estimation (MLE) is usually adopted to optimize the model, and the training objective is defined as:
\begin{equation}
\label{eqn:mle_loss}
\begin{aligned}
L_{mle} &= \sum_{i=1}^{N} \log p(y^i|x^i) \\
&= \sum_{i=1}^{N}\sum_{t=1}^m \log p(y_t^i|y_1^i,...,y_{t-1}^i, x^i),
\end{aligned}
\end{equation}
where $m$ is the length of sentence $y^i$.

Among all the encoder-decoder models, the recently proposed Transformer \cite{transformer} architecture achieves the best translation quality so far. The main difference between Transformer and previous RNNSearch \cite{attention} or ConvS2S \cite{fairseq} is that Transformer relies entirely on self-attention \cite{selfattention} to compute representations of source and target side sentences, without using recurrent or convolutional operations.

\subsection{Training NMT with Reinforcement Learning}

As aforementioned, reinforcement learning (RL) is leveraged to bridge the gap between training and inference of NMT, by directly optimizing the evaluation measure (e.g., BLEU) at training time. Specifically, NMT model can be viewed as an \emph{agent}, which interacts with the \emph{environment} (the previous words $y_{<t}$ and the context vector $z$ available at each step $t$). The parameters of the agent define a \emph{policy}, i.e., a conditional probability $p(y_t|x, y_{<t})$. The agent will pick an \emph{action} , i.e., a candidate word out from the vocabulary, according to the policy. A terminal \emph{reward} is observed once the agent generates a complete sequence $\hat{y}$.  The reward for machine translation is the BLEU \cite{bleu} score, denoted as $R(\hat{y}, y)$, which is defined by comparing the generated $\hat{y}$ with the ground-truth sentence $y$. Note that here the reward $R(\hat{y}, y)$ is the sentence-level reward, i.e., a scalar for each complete sentence $\hat{y}$. The goal of the RL training is to maximize the expected reward:
\begin{equation}
\label{eqn:rl_loss}
\begin{aligned}
L_{rl} &= \sum_{i=1}^{N}E_{\hat{y}\sim p(\hat{y}|x^i)}R(\hat{y}, y^i) \\
&=\sum_{i=1}^{N} \sum_{\hat{y}\in Y} p(\hat{y}|x^i)R(\hat{y}, y^i),
\end{aligned}
\end{equation}
where $Y$ is the space of all candidate translation sentences, which is exponentially large due to the large vocabulary size, making it impossible to exactly maximize $L_{rl}$. In practice, REINFORCE~\cite{reinforce} is usually leveraged to approximate the above expectation via sampling $\hat{y}$ from the policy $p(y|x)$, leading to the objective as maximizing:
\begin{equation}
\label{eqn:reinforce_loss}
\hat{L}_{rl} = \sum_{i=1}^{N}R(\hat{y}^i, y^i), \hat{y}^i \sim p(y|x^i), \forall i \in [N].
\end{equation}

Throughout the paper we will use REINFORCE as our policy optimization method for RL training.

\section{Strategies for RL Training}

Although training NMT with RL can fill in the gap between training objectives and evaluation metrics, it is not easy to successfully put RL training into practice. A key challenge is that RL methods are highly unstable and inefficient, due to the noise in gradient estimation and reward computation. To our best knowledge, currently there is no consensus, or even a systematic study on how to configure different setups for RL training to avoid such problems, especially for training deep NMT models on large scale datasets. We therefore aim to shed light on practical applications of RL for NMT training. For this purpose, we provide a comprehensive review of several important methods to stabilize RL training process in this section.

\subsection{Reward Computation}
\label{subsec:reward}
It is critical to set up appropriate rewards for RL training, i.e., the $R(\hat{y},y)$ in Eqn. (\ref{eqn:reinforce_loss}). There are two important aspects to consider in configuring the reward $R(\hat{y},y)$: how to sample training instance $\hat{y}$ and whether to use reward shaping.

\paragraph{Generate $\hat{y}$}
\label{subsec:sampling}

There are two strategies to sample $\hat{y}$ for computing the BLEU reward $R(\hat{y},y)$. The first one is \emph{beam search} \cite{seq2seq}, it is a breadth-first search method that maintains a ``beam" of the top-$K$ scoring candidates (prefix hypothesis sentences) at each generation step. Then, for each candidate sentence in the beam, $K$ most likely words are appended, resulting in a pool of $K\times K$ new candidates. Out from this pool, the top-$K$ translations with largest probabilities are selected, and the beam search process continues. The second strategy is \emph{multinomial sampling} \cite{sampling}, which produces each word one by one through multinomial sampling over the model's output distribution. Both sampling strategies terminate the expansion of a candidate sentence when an `end of sentence' ($<$EOS$>$) token is met.

The choice of different sampling strategies reflects the \emph{exploration-exploitation} dilemma. Beam search strategy generates more accurate $\hat{y}$ by exploiting the probabilistic space output via current NMT model, while multinomial sampling pays more attention to explore more diverse candidates.

\paragraph{Whether to Use Reward Shaping}

From Eqn. (\ref{eqn:reinforce_loss}) we can see that for the entire sequence $\hat{y}$, there is only one terminal reward $R(\hat{y}, y)$ available for model training. Note that the agent needs to take tens of actions (with the number depending on the length of $\hat{y}$) to generate a complete sentence $\hat{y}$, but only one reward is available for all those actions. Consequently, RL training is inefficient due to the sparsity of rewards, and the model updates each token in the training sentence with the same reward value without distinction. Reward shaping \cite{rewardshaping} is a strategy to overcome this shortcoming. In reward shaping, intermediate reward at each decoding step $t$ is imposed and denoted as $r_t(\hat{y}_t, y)$. \newcite{AC4NMT} sets up the intermediate reward as $r_t(\hat{y}_t, y) = R(\hat{y}_{1...t},y) - R(\hat{y}_{1...{t-1}},y)$, where $R(\hat{y}_{1...t},y)$ is defined as the BLEU score of $\hat{y}_{1...t}$ with respect to $y$. Note that we have $R(\hat{y}, y) = \sum_{t=1}^m r_t(\hat{y}_t, y)$, where $m$ is the length of $\hat{y}$. During RL training, the cumulative reward $\sum_{\tau=t}^m r_{\tau}(\hat{y}_\tau, y)$ is used to update the policy at time step $t$. It is verified that using the shaped reward $r_t$ instead of awarding the whole score $R(\hat{y},y)$ does not change the optimal policy \cite{rewardshaping}.

\subsection{Variance Reduction of Gradient Estimation}
\label{sec:variance}

As mentioned before, the REINFORCE algorithm suffers from high variance in gradient estimation, mainly caused by using single sample $\hat{y}$ to estimate the expectation. To reduce the variance, \newcite{mixer} subtracts an average reward from the returned reward at each time step $t$, and the actual reward used to update the policy is
\begin{equation}
\label{eqn:baseline}
R(\hat{y}, y) - \hat{r}_{t},
\end{equation}
where $\hat{r}_{t}$ is the estimated average reward at step $t$, named as \emph{baseline reward} \cite{weaver2001optimal}. Together with reward shaping, the updated reward becomes $\sum_{\tau=t}^m r_{\tau}(\hat{y}_\tau, y) - \hat{r}_{t}$ at step $t$.

Intuitively speaking, a baseline reward $\hat{r}_{t}$ is established, which either encourages a word choice $\hat{y}_{t}$ if the induced reward $R$ satisfies $R > \hat{r}_{t}$, or discourages it if $R < \hat{r}_{t}$. Here $R$ is either the terminal reward $R(\hat{y}, y)$ or the cumulative reward $\sum_{\tau=t}^m r_{\tau}(\hat{y}_\tau, y)$. Such estimated baseline reward $\hat{r}_{t}$ is designed to decrease the high variance of the gradient estimator.

In practice, the baseline reward $\hat{r}_{t}$ can be obtained through different approaches. For example, one may sample multiple sentences and use the mean terminal reward for these sentences as baseline reward. In our work, we adopt the function learning approach, using simple network (e.g., multi-layer perceptron) to build the learning function, which is the same as used in \cite{mixer,AC4NMT}.

\subsection{Combine MLE and RL Objectives}

The last important strategy we would like to mention is the combination of MLE training objective with RL objective, which is assumed to further stabilize RL training process \cite{gnmt,adversariadialogue,adversarialnmt}.

A simple way is to linearly combine the MLE (Eqn. (\ref{eqn:mle_loss})) and RL (Eqn. (\ref{eqn:reinforce_loss})) objectives as follows:
\begin{equation}
\label{eqn:mle_combine}
L_{com} = \alpha * L_{mle} + (1 - \alpha) * \hat{L}_{rl},
\end{equation}
where $\alpha$ is the hyperparamter controlling the trade-off between MLE and RL objectives. We will empirically evaluate how different values of $\alpha$ impact the final translation accuracy.

\section{RL Training with Monolingual Data}

Previous works typically conduct RL training with only bilingual data for NMT. Monolingual data has been proved to be able to significantly improve the performance of NMT systems~\cite{backtranslation, duallearning, chengyong}. It remains an open problem whether it is possible to combine the benefits of RL training and monolingual data such that even more competitive results can be obtained. In this section we provide several solutions for combination and will study them in next section. Note that all the settings discussed in this section are semi-supervised learning, i.e., both bilingual and monolingual data are available.

\subsection{With Source-Side Monolingual Data}
\label{subsec:src_mono}

We first provide a solution to RL training with source-side monolingual data.  As shown in Eqn. (\ref{eqn:reinforce_loss}), in RL training we need to calculate the reward signal $R(\hat{y}, y)$ for each generated sentence $\hat{y}$, and therefore the reference sentence $y$ seems to be a must-have, which unfortunately is missing for source-side monolingual data.

We tackle this challenge via generating pseudo target reference $y$ by bootstrapping with the model itself. Apparently, for the source-side monolingual data, the pseudo target reference $y$ should have good translation quality. Therefore, for each source-side monolingual sentence, we use the NMT model trained from the bilingual data to beam search a target sentence and treat it as the pseudo target reference $y$.  Afterwards $\hat{y}$ is obtained via multinomial sampling to calculate the reward. Although multinomial sampling is usually not as good as sampling via beam search, the combination of beam search (to get the pseudo target reference sentence) and the multinomial sampling (to generate the action sequence of the agent) achieves good exploration-exploitation trade-off, since the pseudo target reference exploits the accuracy of current NMT model while $\hat{y}$ achieves better exploration.

\subsection{With Target-Side Monolingual Data}
\label{subsec:tgt_mono}

For a target-side monolingual sentence, its source sentence $x$ is missing, and consequently $\hat{y}$ is unavailable since it is sampled based on $x$.  We tackle this challenge via back translation~\cite{backtranslation}. We first train a reverse NMT model from the target language to the source language with bilingual data. For each target-side monolingual sentence, using the reverse NMT model, we back translate it to get its pseudo source sentence $x$. We then pair the target monolingual data and its back-translated sentence as a pseudo bilingual sentence pair, which can be used for RL training in the same way as the genuine bilingual sentence pairs.

\subsection{With both Source-Side and Target-Side Monolingual Data}
\label{sec:cross_mono}

A natural extension of previous discussions is to combine both the source-side and target-side monolingual data for RL training. We consider two combinations, the \emph{sequential} method and the \emph{unified} method.  The former one sequentially leverages the source-side and target-side monolingual data for RL training. Specifically, we first train an MLE model using the bilingual data and source-side (or target-side) monolingual data; based on this MLE model, we then use REINFORCE for training with target-side (or source-side) monolingual data. For \emph{unified} approach, we pack the paired data out from three domains together: the genuine bilingual data, the source monolingual data with its pseudo target references (introduced in subsection~\ref{subsec:src_mono}), and the target monolingual data with its back-translated samples (introduced in subsection~\ref{subsec:tgt_mono}). Then we treat the combined data as normal bilingual data on which the NMT model is trained via MLE or RL principles. Our goal is to investigate the model performance with different training data and find the best recipe of how to use these data in RL training. More details are introduced in next section.

\section{Experiments}

In this section, we provide a systematic study on aforementioned RL training strategies and the solutions of leveraging monolingual data. The RL training strategies are evaluated on bilingual datasets from three translation tasks, WMT14 English-German (En-De), WMT17 English-Chinese (En-Zh) and WMT17 Chinese-English (Zh-En), and we further conduct the experiments to leverage monolingual data in WMT17 Zh-En translation.

\subsection{Experimental Settings}

For the bilingual datasets,  WMT17~\cite{WMT17} En-Zh \footnote{\url{http://www.statmt.org/wmt17/translation-task.html}} and WMT17 Zh-En use the same dataset, which contains about $24M$ sentences pairs, including CWMT Corpus 2017 and UN Parallel Corpus V1.0.  The Jieba\footnote{\url{https://github.com/fxsjy/jieba}} segmenter is used to perform Chinese word segmentation. We use byte pair encoding (BPE) \cite{bpe} to preprocess the source and target sentences, forming source-side and target-side dictionary with $40,000$ and $37,000$ types, respectively. We use the \emph{newsdev2017} as the dev set and \emph{newstest2017} as the test set. For the WMT14 En-De dataset, it contains about $4.5M$ training pairs, \emph{newstest2012} and \emph{newstest2013} are concatenated as the dev set and \emph{newstest2014} acts as test set. Same as \cite{transformer}, we also perform BPE to process the En-De dataset, the shared source-target vocabulary contains about $37,000$ tokens.

For the monolingual dataset on Zh-En translation task, similar to \cite{UE_wmt17}, the Chinese monolingual data comes from LDC Chinese Gigaword (4th edition) and the English monolingual data comes from News Crawl 2016 articles. After  preprocessing (e.g., language detection and filtering sentences with more than $80$ words), we keep $4M$ Chinese sentences and $7M$ English sentences.

We adopt the Transformer model with \emph{transformer\_big} setting as defined in \cite{transformer} for Zh-En and En-Zh translations, which achieves SOTA translation quality in several other datasets. For En-De translation, we utilize the \emph{transformer\_base\_v1} setting. These settings are exactly same as used in the original paper, except we set the \emph{layer\_prepostprocess\_dropout} for Zh-En and En-Zh translation to be $0.05$. The optimizer used for MLE training is Adam \cite{adam} with initial learning rate is $0.1$, and we follow the same learning rate schedule in \cite{transformer}. During training, roughly $4,096$ source tokens and $4,096$ target tokens are paired in one mini batch. Each model is trained using $8$ NVIDIA Tesla M40 GPUs. For RL training, the model is initialized with parameters of the MLE model (trained with only bilingual data), and we continue training it with learning rate $0.0001$. Same as \cite{AC4NMT}, to calculate the BLEU reward, we start all n-gram counts from $1$ instead of $0$ and multiply the resulting score by the length of the target reference sentence. For inference, we use beam search with width $6$. We run each setting for at least 5 times and report the averaged case sensitive BLEU scores\footnote{Calculated by Sacr\'eBLEU toolkit, which produces exactly the same evaluation result as that in WMT17 Zh-En campaign. \url{https://github.com/awslabs/sockeye/tree/master/contrib/sacrebleu}} \cite{bleu} on test set. The test set BLEU is chosen via the best configuration based on the validation set.


\subsection{Results of of RL Training Strategies}

We first evaluate different strategies for RL training, based only on bilingual datasets from previously introduced three translation tasks.

\paragraph{Reward Computation}
As reviewed in subsection~\ref{subsec:reward}, for reward computation, we need to consider how to sample $\hat{y}$ and whether to use reward shaping.

The results are shown in Table \ref{tab:sample_reward}, where ``RL" stands for RL training with the REINFORCE algorithm. We also report the performance of the pre-trained NMT model with the MLE loss. From the table, an interesting finding is that $\hat{y}$ sampled via beam search strategy is worse than that by multinomial sampling, with a gap of roughly $0.2$-$0.3$ BLEU points on the test set (with significant test score $\rho< 0.05$). We therefore conjecture that exploration is more important than exploitation in reward computing: multinomial sampling brings more data diversity to the training of NMT model, while sentences generated by beam search are usually very similar to each other.
Furthermore, we find that there is no big difference between the leverage of reward shaping or terminal reward, with only slightly better performance of reward shaping. We therefore use multinomial sampling and reward shaping in later experiments.

\begin{table}[t!]
\centering
\small
\begin{tabular}{c | c c c}
\hline
\textbf{Training Strategy} & \textbf{En-De} & \textbf{En-Zh}& \textbf{Zh-En}\\[0.25ex]
\hline \hline
MLE & 27.02 & 34.12 & 24.29\\[0.25ex]
RL (beam + terminal) & 27.06 & 34.25 & 24.42  \\[0.25ex]
RL (multinomial + terminal) & \textbf{27.22} &\textbf{34.46} & \textbf{24.70}  \\[0.25ex]
RL (beam + shaping) & 27.04 & 34.28 & 24.47  \\[0.25ex]
RL (multinomial + shaping) & \textbf{27.23} & \textbf{34.47} & \textbf{24.72}  \\[0.25ex]
\hline
\end{tabular}
\caption{Results of different strategies for reward computation. `beam' refers to `beam search¡¯ and `multinomial' to `multinomial sampling'. While generating $\hat{y}$ through beam search, we use width 4. `shaping' refers to using reward shaping and `terminal' refers not.}
\label{tab:sample_reward}
\end{table}


\paragraph{Variance Reduction of Gradient Estimation}

Next we evaluate the strategies for reducing variance of gradient estimation (see section\ref{sec:variance}). We want to know whether the \emph{baseline reward} is necessary. To compute the baseline reward, similar to \cite{mixer,AC4NMT}, we build a two-layer MLP regressor with Relu \cite{relu} activation units. The function takes the hidden states from decoder as input, and the parameters of the regressor are trained to minimize the mean squared loss of Eqn. (\ref{eqn:baseline}). We first pre-train the baseline function for $20k$ steps/mini-batches, and then jointly train NMT model (with RL) and the baseline reward function.

\begin{table}[t!]
\small
\centering
\begin{tabular}{c | c c c}
\hline
\textbf{Training Strategy} & \textbf{En-De}& \textbf{En-Zh}& \textbf{Zh-En}\\[0.25ex]
\hline \hline
RL & 27.23 & 34.47 & 24.72  \\[0.25ex]
RL (baseline function) & 27.25 & 34.43 & 24.73  \\[0.25ex]
\hline
\end{tabular}
\caption{Results of variance reduction of gradient estimation.}
\label{tab:baseline}
\end{table}

Table \ref{tab:baseline} shows that the learning of baseline reward does not help RL training. This contradicts with previous observations~\cite{mixer}, and seems to suggest that the variance of gradient estimation in NMT is not as large as we expected. The reason might be that the probability mass on the target-side language space induced by the NMT model is highly concentrated, making the sampled $\hat{y}$ representative enough in terms of estimating the expectation. Therefore, for the economic perspective, it is not necessary to add the additional steps of using baseline reward on RL training for NMT.

\paragraph{Combine MLE and RL Objectives}

\begin{figure}
  \centering
  \includegraphics[width=1.1\linewidth]{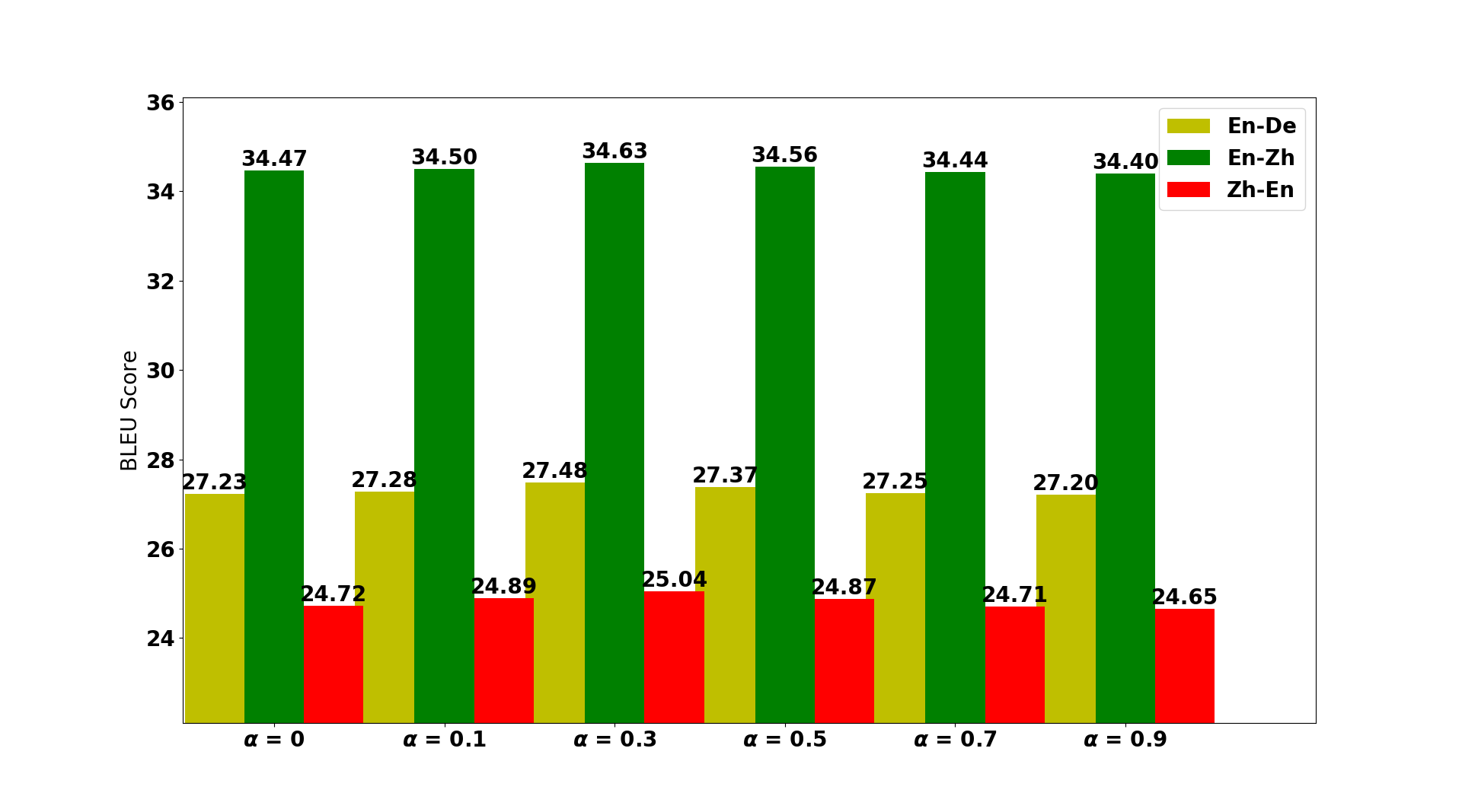}
  \caption{Results of different weights $\alpha$ to combine MLE and RL objectives.}
  \label{fig:mle_combine}
\end{figure}

As shown in Eqn. (\ref{eqn:mle_combine}), the hyperparameter $\alpha$ controls the trade-off between MLE and RL objectives. For  comparison, we set $\alpha$ to be [0, 0.1, 0.3, 0.5, 0.7, 0.9] in our experiments. The results are presented in Figure \ref{fig:mle_combine}.

The results show that combining the MLE objective with the RL objective achieves better performance ($27.48$ for En-De, $34.63$ for En-Zh and $25.04$ for Zh-En with $\alpha=0.3$). This indicates that MLE objective is helpful to stabilize the training and improve the model performance, as we expected. However, further increasing $\alpha$ does not bring more gain. The best trade-off between MLE and RL objectives in our experiment is $\alpha=0.3$. Therefore, we set $\alpha = 0.3$ in the following experiments.


\subsection{Results of RL Training with Monolingual Data}

In this subsection, we report the results on both valid and test set of RL training using bilingual and monolingual data in Zh-En translation. From Table~\ref{tab:source_mono} to Table \ref{tab:combine_mono}, ``RL" denotes the model trained with RL using multinomial sampling, reward shaping, no baseline reward, and combined objective, based on the observations in the last subsection. ``B" denotes bilingual data, ``Ms" denotes source-side monolingual data and ``Mt" denotes target-side monolingual data, ``\&" denotes data combination.

\paragraph{With Source-Side Monolingual Data}

As discussed before, we use beam search with beam width 4 to sample the pseudo target sentence $y$ for each monolingual sentence $x$. We consider several settings for RL training: 1) only source-side monolingual data; 2) the combination of bilingual and source-side monolingual data. We first train an MLE model using the augmented dataset combining the genuine bilingual data with the pseudo bilingual data generated from the monolingual data, and then perform RL training on this combined dataset. The results are shown in Table \ref{tab:source_mono}.

\begin{table}[t!]
\small
\centering
\begin{tabular}{c | c c}
\hline
\textbf{[Data] (Objective)} & \textbf{Valid}& \textbf{Test}\\[0.25ex]
\hline \hline
[B] (MLE) & 22.32 & 24.29 \\[0.25ex]
[B] (MLE) + [B] (RL) & 22.87 & 25.04  \\[0.25ex]
[B] (MLE) + [Ms] (RL) & \textbf{23.03} & \textbf{25.22}  \\[0.25ex]
\hline
[B \& Ms] (MLE) & 24.31 & 25.31 \\[0.25ex]
[B \& Ms] (MLE) + [B \& Ms] (RL) & 24.58 & 25.60 \\[0.25ex]
\hline
\end{tabular}
\caption{Results with source monolingual data. ``B" denotes bilingual data, ``Ms" denotes source-side monolingual data, ``\&" denotes data combination.}
\label{tab:source_mono}
\end{table}

\paragraph{With Target-Side Monolingual Data}


For target-side monolingual data, we first pre-train a translation model from English to Chinese \footnote{The BLEU score of the En-Zh model is $34.12$.}, and use it to back translate target-side monolingual sentence $y$ to get pseudo source sentence $x$. Similarly, we consider several settings for RL training: 1) only target-side monolingual data; 2) the combination of bilingual data and target-side monolingual data. We train an MLE model using both the genuine and the generated pseudo bilingual data, and then perform RL training on this data. The results are presented in Table \ref{tab:target_mono}.

\begin{table}[t!]
\small
\centering
\begin{tabular}{c | c c}
\hline
\textbf{[Data] (Objective)} & \textbf{Valid}& \textbf{Test}\\[0.25ex]
\hline \hline
[B] (MLE) & 22.32 & 24.29 \\[0.25ex]
[B] (MLE) + [B] (RL) & 22.87 & 25.04  \\[0.25ex]
[B] (MLE) + [Mt] (RL) & \textbf{22.96} & \textbf{25.15}  \\[0.25ex]
\hline
[B \& Mt] (MLE) & 24.14 & 25.24 \\[0.25ex]
[B \& Mt] (MLE) + [B \& Mt] (RL) & 24.41 & 25.58 \\[0.25ex]
\hline
\end{tabular}
\caption{Results with target monolingual data. ``B" denotes bilingual data, ``Mt" denotes target-side monolingual data, ``\&" denotes data combination.}
\label{tab:target_mono}
\end{table}

From Table \ref{tab:source_mono} and ~\ref{tab:target_mono}, we have several observations. First, monolingual data helps RL training, improving BLEU score from $25.04$ to $25.22$ ($\rho<$ 0.05) in Table \ref{tab:source_mono}. Second, when we only add monolingual data for RL training, the model achieves similar performance compared to MLE training with bilingual and monolingual data (e.g., $25.15$ vs. $25.24$ ($\rho<$ 0.05) in Table \ref{tab:target_mono}).

\paragraph{With both Source-Side and Target-Side Monolingual Data}

We have two approaches to use both source-side and target-side monolingual data, as described in subsection \ref{sec:cross_mono}. The results are reported in Table \ref{tab:cross_mono} and Table \ref{tab:combine_mono}.

\begin{table}[t!]
\small
\centering
\begin{tabular}{c | c c}
\hline
\textbf{[Data] (Objective)} & \textbf{Valid}& \textbf{Test}\\[0.25ex]
\hline \hline
[B \& Ms] (MLE) & 24.31 & 25.31 \\[0.25ex]
[B \& Ms] (MLE) + [B \& Ms] (RL) & 24.58 & 25.60  \\[0.25ex]
[B \& Ms] (MLE) + [Mt] (RL) & \textbf{24.61} & \textbf{25.72}  \\[0.25ex]
\hline
[B \& Mt] (MLE) & 24.14 & 25.24 \\[0.25ex]
[B \& Mt] (MLE) + [B \& Mt] (RL) & 24.41 & 25.58 \\[0.25ex]
[B \& Mt] (MLE) + [Ms] (RL) & \textbf{24.75} & \textbf{25.92}  \\[0.25ex]
\hline
\end{tabular}
\caption{Results of \emph{sequential} approach for monolingual data. ``B" denotes bilingual data, ``Ms" denotes source-side monolingual data and ``Mt" denotes target-side monolingual data, ``\&" denotes data combination.}
\label{tab:cross_mono}
\end{table}

\begin{table}[t!]
\small
\centering
\begin{tabular}{c | c c}
\hline
\textbf{[Data] (Objective)} & \textbf{Valid}& \textbf{Test}\\[0.25ex]
\hline \hline
[B \& Ms \& Mt] (MLE) & 25.58 & 26.13  \\[0.25ex]
+ [B \& Ms \& Mt] (RL) & \textbf{25.90} & \textbf{26.73}  \\[0.25ex]
\hline
\end{tabular}
\caption{Results of \emph{unified} approach for monolingual data. ``+" means to initialize the RL model using above MLE model, which is trained on the combination of bilingual data, source-side monolingual data and target-side monolingual data.}
\label{tab:combine_mono}
\vspace{-5pt}
\end{table}

\begin{table*}[!t]
\small
\centering
\begin{minipage}{13.5cm}
\begin{tabular}{|l | l | c | c | c}
\hline
System & Architecture  & BLEU \\
\hline \hline
\multicolumn{4}{c}{{\em Existing end-to-end NMT systems}} \\
\hline
\newcite{transformer} & Transformer & 24.29\\
\newcite{backtranslation} & Transformer + Target Monolingual Data (i.e., back translation) & 25.24 \\
SougouKnowing & Stacked LSTM model + Reranking & 24.00\\
SougouKnowing-ensemble & Stacked LSTM model + Reranking + Ensemble & 26.40\\
\hline
\multicolumn{4}{c}{{\em Our end-to-end NMT}} \\
\hline
\multirow{3}{*{\em this work}}{} & Transformer + \emph{RL} & 25.04 \\
& Transformer + Source Monolingual Data & 25.31 \\
& Transformer + Source Monolingual Data + \emph{RL} & 25.60 \\
& Transformer + Target Monolingual Data & 25.24 \\
& Transformer + Target Monolingual Data + \emph{RL} & 25.58 \\
& Transformer + Source \& Target Monolingual Data & 26.13 \\
& Transformer + Source \& Target Monolingual Data + \emph{RL} & \textbf{26.73} \\
\hline
\end{tabular}
\end{minipage}
\caption{Comparisons of different competitive end-to-end NMT systems. SougouKnowing results come from \url{http://matrix.statmt.org/matrix/systems_list/1878}.}
\label{tab:result_all}
\vspace{-7pt}
\end{table*}

From Table \ref{tab:cross_mono}, we can observe that the \emph{sequential} training of monolingual data can benefit the model performance. Taking the last three rows as an example, the BLEU score of the MLE model trained on the combination of bilingual data and target-side monolingual data is $25.24$; based on this model, RL training using the source-side monolingual data further improves the model performance by $0.7$ ($\rho<$ 0.01) BLEU points. From Table \ref{tab:combine_mono}, we can observe on top of a quite strong MLE baseline ($26.13$), through the \emph{unified} RL training, we can still improve the test set by $0.6$ points to $26.73$ ($\rho<$ 0.01), which shows the effectiveness of combining source/target monolingual data and reinforcement learning.

\subsection{Comparison with Other Models}

At last, as a summary of our empirical results, we compare several representataive end-to-end NMT systems to our work in Table \ref{tab:result_all}, which includes the Transformer \cite{transformer} model, with/without back-translation \cite{backtranslation} and the best NMT system in WMT17 Chinese-English translation challenge\footnote{\url{http://matrix.statmt.org/matrix/systems_list/1878}} (SougouKnowing-ensemble). The results clearly show that after combing both source-side and target-side monolingual data with RL training, we obtain the state-of-the-art BLEU score $26.73$, even surpassing the best ensemble model in WMT17 Zh-En translation challenge.

\section{Related Work}

Our work is mainly related with the literature of using reinforcement learning to directly optimize the evaluation measure for neural machine translation. Several representative works are \cite{mixer,MRT,AC4NMT}. In \cite{mixer}, the authors propose to train a neural translation model with the objective gradually shifting from maximizing token-level likelihood to optimizing the sentence-level BLEU score. \newcite{MRT} proposes to adopt minimum risk training \cite{mbr} to minimize the task specific expected loss (i.e., induced by BLEU score) on NMT training data. Instead of the REINFORCE~\cite{reinforce} algorithm used in the above two works, \newcite{AC4NMT} further optimizes the policy by actor-critic algorithm. \newcite{gnmt} introduces a simple RL based method to optimize the stacked LSTM model for NMT, achieving better BLEU scores on English-French translation but not on English-German. \newcite{structureloss} presents a comparative study of several classical structural prediction losses for NMT model, which also includes sequence-level loss but not exactly the same as RL.

Our work is also related with the research works that leverage monolingual data for improving NMT models \cite{semi,backtranslation,wang2018dual,duallearning,chengyong}. \newcite{semi} exploits the source-side monolingual data in NMT. \newcite{backtranslation} proposes back-translation method to leverage target-side monolingual data for NMT. \newcite{duallearning} formulates the machine translation as a communication game, which leverages the power of two directional translation models and source/target monolingual data. \newcite{chengyong} proposes a similar semi-supervised approach. However, none of these works have explored the power of monolingual data in the context of training NMT model with reinforcement learning.

\section{Conclusion}

In this work, we presented a study of how to effectively train NMT models using reinforcement learning. Different RL strategies were evaluated in German-English, English-Chinese and Chinese-English translation tasks on large-scale bilingual datasets. We found that (1) multinomial sampling is better than beam search, (2) several previous tricks such as reward shaping and baseline reward does not make significant difference, and (3) the combination of the MLE and RL objectives is important. In addition, we explored the source/target monolingual data for RL training. By combing the power of RL and monolingual data, we achieve the state-of-the-art BLEU score on WMT17 Chinese-English translation task. We hope that our study and results can benefit the community and bring some insights on how to train deep NMT models with reinforcement learning and big data.

\bibliography{emnlp2018}
\bibliographystyle{acl_natbib_nourl}

\end{document}